\documentclass[11pt,a4paper]{article}
\usepackage[hyperref]{emnlp-ijcnlp-2019}
\usepackage{times}
\usepackage{latexsym}

\usepackage{url}

\PassOptionsToPackage{usenames,dvipsnames,table}{xcolor}
\relax

\usepackage{helvet}  
\usepackage{courier}  
\usepackage{graphicx}  
\usepackage{quoting}
\usepackage[utf8]{inputenc} 
\usepackage[T1]{fontenc}    
\usepackage{booktabs}       
\usepackage{amsfonts}       
\usepackage{nicefrac}       
\usepackage{microtype}      
\usepackage{amssymb}
\usepackage{amsmath}
\usepackage{multirow}
\usepackage{amsfonts}
\usepackage{booktabs}
\usepackage{rotating}
\usepackage{subfig}
\usepackage{array}
\usepackage{booktabs}
\usepackage{amsfonts}
\usepackage{enumitem}
\usepackage{makecell}
\usepackage{xcolor}
\usepackage[toc,page]{appendix}
\usepackage{amsmath}
\usepackage{xspace}
\usepackage{mdwlist}
\usepackage{graphics}
\usepackage{color}
\usepackage{rotating}
\usepackage{booktabs}
\usepackage{epsfig}
\usepackage{alltt}
\usepackage{moreverb}
\usepackage{fancyvrb}
\usepackage{colortbl}      
\usepackage{xspace}
\usepackage{relsize}
\usepackage{array}
\usepackage{amsmath}
\usepackage{amsfonts}
\usepackage{txfonts}
\usepackage{caption}
\usepackage{rotating}
\usepackage[]{algorithm2e}
\usepackage{multirow}
\usepackage{booktabs}
\usepackage{float}
\usepackage{array,etoolbox}
\usepackage{subfig}
\usepackage{arydshln}
\usepackage{setspace}
\usepackage{float}
\usepackage{mathtools}
\usepackage{bm}
\usepackage{xcolor}
\usepackage[toc,page]{appendix}
\usepackage{soul}
\usepackage[draft]{pgf}
\usepackage{makecell}
\usepackage[normalem]{ulem}
\usepackage{amssymb}
\usepackage{wasysym,textcomp}
\usepackage[toc,page]{appendix}
\usepackage{wrapfig}

\usepackage{tabularx}

\definecolor{Red}{rgb}{1,0,0}
\definecolor{Green}{rgb}{0,0.7,0}
\definecolor{Blue}{rgb}{0,0,1}
\definecolor{Red}{rgb}{0.6,0,0}
\definecolor{Orange}{rgb}{1,0.5,0}

\usepackage{etoolbox}
\newtoggle{draft}
\toggletrue{draft}
\iftoggle{draft}{

}{
\newcommand{\todo}[1]{}
}

\newcommand{\com}[1]{}

\mathchardef\mhyphen="2D
\newenvironment{des}{                 
     \parskip 0cm \begin{list}{}{\parsep 0cm \itemsep 0cm \topsep 0cm}}{
       \end{list}} 

\newcommand{\wiqabert}[1]{\textsc{wiqa-bert}}

\newcommand{\ourdata}{\textsc{WIQA}}

\newenvironment{myquote}{                   
  \parskip 0mm \begin{quoting}[vskip=0mm,leftmargin=2mm]}{
\end{quoting}}

\newenvironment{mycentering}
 {\parskip=0pt\par\nopagebreak\centering}
 {\par\noindent\ignorespacesafterend}

\aclfinalcopy 

\title{\ourdata: A dataset for "What if..." reasoning over procedural text}
\author{
  Niket Tandon\textsuperscript{*}, Bhavana Dalvi Mishra\thanks{\textsuperscript{*}Niket Tandon and Bhavana Dalvi Mishra contributed equally to this work.} , Keisuke Sakaguchi, \\
  {\bf Antoine Bosselut, Peter Clark} \vspace{1mm} \\
        Allen Institute for Artificial Intelligence, Seattle, WA \\
        {\tt \{nikett,bhavanad,keisukes,antoineb,peterc\}@allenai.org}
}

\date{}

\hypersetup{final}
\begin{document}
\maketitle

\begin{abstract}
We introduce \ourdata, the first large-scale dataset of "What if..."
questions over procedural text. \ourdata~contains three parts:
a collection of paragraphs each describing a process, e.g., beach erosion;
a set of crowdsourced {\it influence graphs} for each paragraph, describing
how one change affects another;
and a large (40k) collection of
"What if...?" multiple-choice questions derived from the graphs.
For example, given a paragraph about beach erosion, would
stormy weather result in more or less erosion (or have no effect)?
The task is to answer the questions, given their associated paragraph.
\ourdata~contains
three kinds of questions: perturbations to steps mentioned in
the paragraph; external (out-of-paragraph) perturbations requiring
commonsense knowledge; and irrelevant (no effect) perturbations.
We find that state-of-the-art models achieve 73.8\% accuracy,
well below the human performance of 96.3\%. We analyze the challenges,
in particular tracking chains of influences, and present
the dataset as an open challenge to the community.
\end{abstract}

\begin{figure}[t]
{\small 
{\bf Procedural Text} (simplified): \vspace{1mm} \\
\centerline{
\fbox{%
    \parbox{0.34\textwidth}{%
      \underline{\bf Erosion by the ocean:} \\
1. Wind creates waves in the ocean. \\
2. The waves wash onto the beaches. \\
3. The waves hit rocks on the beach. \\
4. Tiny parts of the rock break off. \\
5. The rocks become smaller.
    } 
} 
} 
\vspace{-3mm}
\begin{center}
{\includegraphics[width=1.08\columnwidth]{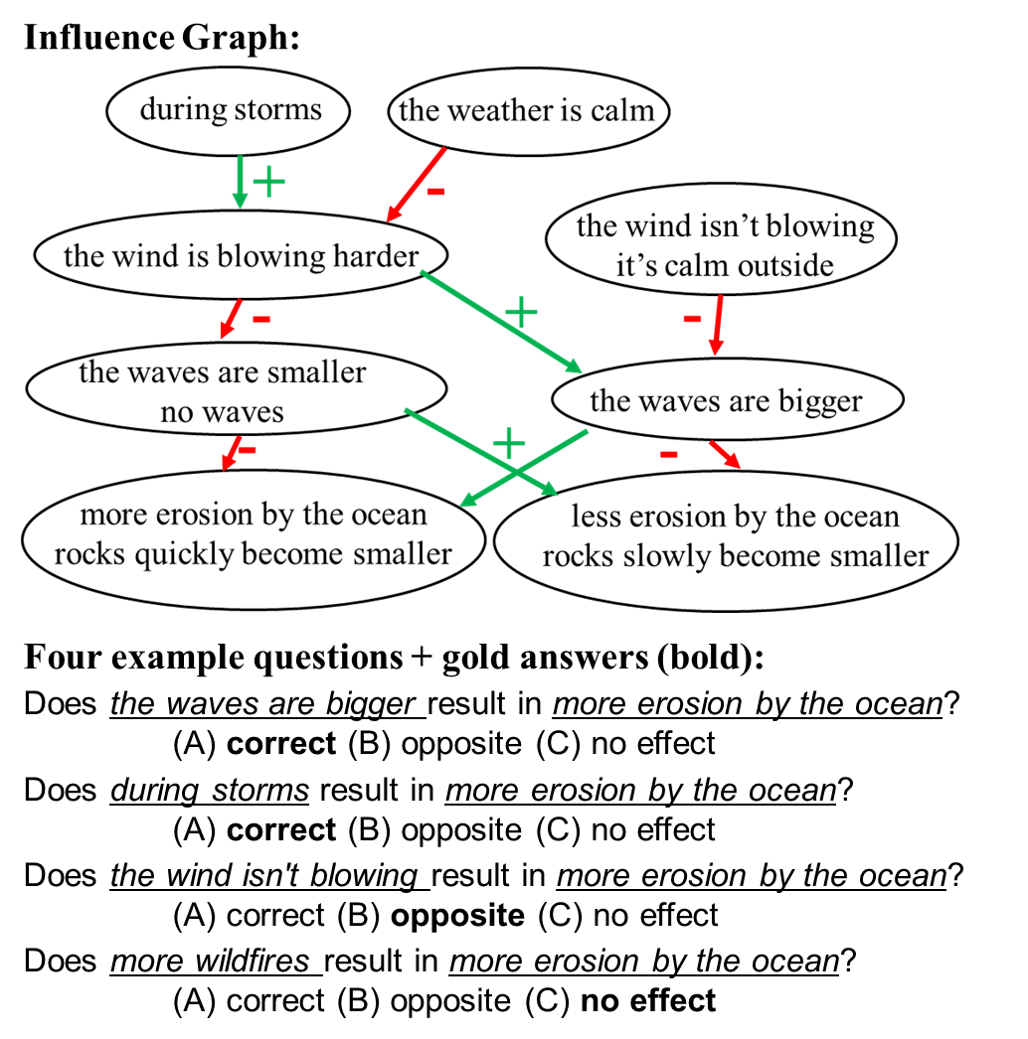}}
\end{center}
}
\vspace{-3mm}
\caption{
  \ourdata~contains procedural paragraphs, crowdsourced influence graphs associated with them,
  and a large collection of ``Does {\it changeX} result in {\it changeY}?'' (what-if) questions,
  derived from the graphs.
  \label{fig:influence_graph}}
\vspace{-8mm}
\end{figure}

\section{Introduction}

Procedural text is common in language, but challenging to comprehend because it describes a dynamically
changing world. While recent systems for procedural text comprehension can answer questions about what events happen,
e.g., \cite{npn,Henaff2016TrackingTW,propara-naacl18}, 
the extent to which they understand the influences {\it between} those events remains unclear.

One important test of understanding is to predict what would happen if 
a process was {\it perturbed} in some way, requiring understanding and tracing the chain 
of influences through a paragraph. However, to date there is no dataset
available to help develop this capability. We aim to fill this gap with
WIQA\footnote{The dataset is available at http://data.allenai.org/wiqa/}, the first large-scale dataset testing "What if..." reasoning over
procedural text. 

WIQA contains 40.7K questions, for 379 process paragraphs.
To efficiently create the questions, crowdworkers created
2107 {\it influence graphs} (IGs) for the paragraphs, describing how one perturbation positively or
negatively influences another (Figure~\ref{fig:influence_graph}).
Questions were then derived from paths in the graphs, each asking how the
change described in one node affects another. Each question
is a templated, multiple choice (MC) question
of the form {\it Does changeX result in changeY? (A) Correct (B) Opposite (C) No effect},
where {\it Opposite} indicates a negative influence between {\it changeX} and {\it changeY}.
To bound the task, perturbations are typically qualitative (e.g., ``the wind is blowing harder''),
and possible effects are restricted to changes to entities and events mentioned in the paragraph
(e.g., ``the waves are bigger''). Perturbations themselves include
in-paragraph, out-of-paragraph, and irrelevant (no effect) changes.
The WIQA task is to answer the questions, given the paragraph (but not the IG).

We first describe the task and how the dataset was constructed, and then
present results from baselines and strong BERT-based models.
We find that the best model is still 23\% behind human performance and the gap further widens with indirect and out-of-paragraph effects,
illustrating that the dataset is hard. We present a detailed
analysis showing WIQA is rich in linguistic and semantic phenomena.
Our contributions are: (1) the new dataset (2) performance measures
and an analysis of its challenges, to support research on 
counterfactual, textual reasoning over procedural text.

\vspace{-1mm}
\section{Related Work}
\vspace{-1mm}
While there are several NLP datasets now available for
procedural text understanding, e.g., \cite{Kiddon2016GloballyCT,propara-naacl18,weston2015towards},
these have all targeted the task of tracking entity states
throughout the text. \ourdata~takes the next step of asking
how states might {\it change} if a perturbation was introduced.

Predicting the effects of qualitative change has been studied in
the qualitative reasoning (QR) community, but primarily using
formal models \cite{Forbus1984QualitativePT,weld2013readings}.
Similarly, counterfactual reasoning has been studied in
the logic community \cite{lewis2013counterfactuals},
but again using formal frameworks. In contrast, \ourdata~ treats
the task as a mixture of reading comprehension and
commonsense reasoning, creating a new NLP challenge.

\vspace{-1mm}
\section{Dataset Construction \label{dataset}}
\vspace{-1mm}

To efficiently generate questions, we first asked crowdworkers to create
{\it influence graphs} (IGs) for each paragraph. We then create questions
from the IGs using paths in the IGs. We now describe this process.

\label{sec:problem}

\paragraph{Influence Graphs} 
\label{defn:influence_graph}
An influence graph $\mathcal{G}(V,E)$ for a procedural text $T$ is an unweighted directed graph.
Each vertex $v_i$ is labeled with one or more text strings, each describing a change to
the original conditions described or assumed in $T$, such that all those changes
have the same influence on a connected node $v_j$. Each edge is labeled with a 
{\it polarity}, {\bf +} or {\bf -},
indicating whether the influence is positive (causes/increases) or negative (prevents/ reduces).


Indirect effects can be found by traversing $\mathcal{G}$.

\begin{figure}[t]
  \includegraphics[width=0.5\textwidth]{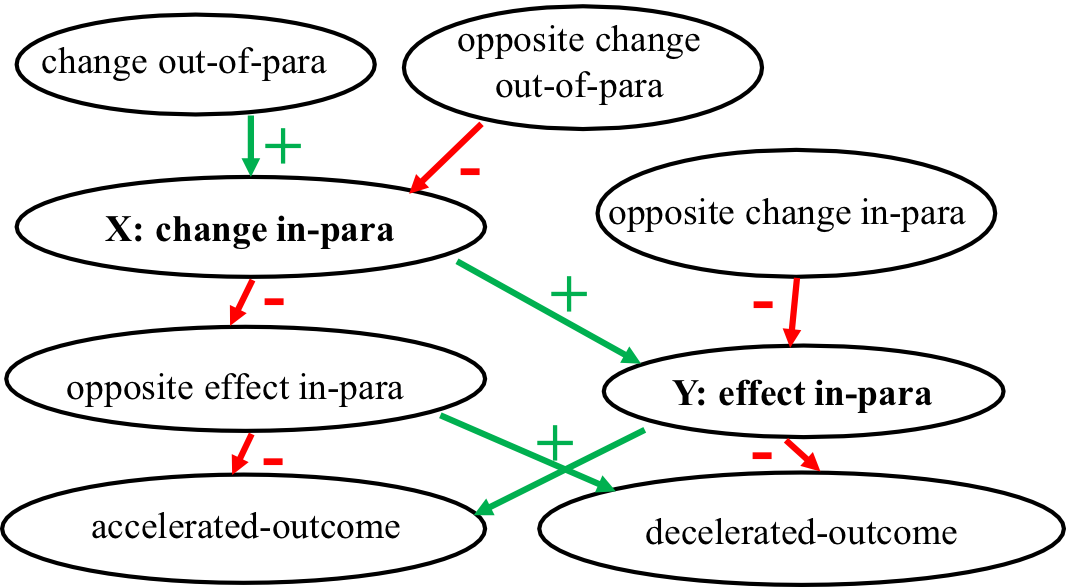}
  \caption{The template used to acquire influence graphs} 
  \label{fig:consistency}
\end{figure}

\noindent
It is useful to distinguish two kinds of nodes: 

\noindent
1. {\bf Out-of-para nodes:} denoting events or changes to entities/events not mentioned in the paragraph, e.g, ``during storms'' in Figure~\ref{fig:influence_graph}.

\noindent
2. {\bf In-para nodes:} denoting events or changes to entities/events mentioned in the paragraph,  e.g, ``the wind is blowing harder'' in Figure~\ref{fig:influence_graph}.

\paragraph{Acquiring influence graphs}


For a source of paragraphs, we used the 377 training set paragraphs from the ProPara dataset \cite{propara-emnlp18}. (Multiple) influence graphs were then crowdsourced for each. To do this, we use
an {\it influence graph template}, shown in Figure~\ref{fig:consistency}. Workers were asked to populate
this (hidden) template using a sequence of five questions, where the later questions
were automatically constructed from their answers to the earlier questions.
The first question asks the worker to supply an X and Y in: ``If [X] occurs, it will have the
intermediate effect [Y] resulting in {\it accelerated\_outcome}'' (where the {\it accelerated outcome} phrase was pre-authored for each paragraph).
For X and Y, workers were asked to describe a change in some property/phenomonon mentioned in the paragraph,
e.g., if a paragraph sentence $x_{i}$ is ``Wind creates waves.'', they may author an X saying
``the wind is blowing harder'' (Figure~\ref{fig:influence_graph}).
(The alignment of X and $x_{i}$,
and whether X describes an increase or decrease of $x_{i}$, denoted by $d_{X} \in \{+,-\}$,
was also recorded.).
This fills X and Y in Figure~\ref{fig:consistency}. Similar questions populate the remaining nodes (see Appendix).
2107 influence graphs were collected in this way.

\paragraph{Generating Questions from Graphs}

Each path in a graph forms a {\it ``change$\rightarrow$effect?''} question, whose answer is either ``correct'' or
``opposite'' depending on the product of the polarities of the traversed edges. Questions are labeled with
the number of edges traversed (1 = "1-hop", etc.). 
We also distinguish {\it in-para} and {\it out-of-para} questions depending on the type of
node they originated from. We then created a third
category of question, whose answer is ``no effect'', by selecting out-of-para changes from {\it other} paragraphs and asking for
their effect on nodes in the current graph. Occasionally these changes did affect the selected node, resulting in an erroneous label of "no effect", but this was rare (and such cases were removed from the test partition, as we now describe).

Using a separate crowdsourcing task, questions in the test set were filtered
to improve the test set quality. First, five workers independently answered each question, given the paragraph. The
inter-annotator agreement between workers, using Krippendorff's alpha, was moderately high (0.6).
We then retained only questions with majority agreement
(i.e., at least 3 out of 5 workers agreed), resulting in 88\% of questions being retained.

\paragraph{Balancing the Dataset}
From the (many) questions thus generated, we (randomly) selected a subset that
approximately balanced the numbers of (a) in-para, out-of-para, and no-effect questions, and
(b) questions with each answer (correct, opposite, no-effect), resulting in 40,695 questions. Train, dev, and test partitions do not share paragraphs about the same topic.
Statistics are shown in Table~\ref{table:stats_main}. 

\paragraph{Explanations}

As each question is derived from a path in an IG, we can also generate {\it explanations} for
each answer using that path. Although explanation is not part of the \ourdata~task,
we create an explanation database to support a possible explanation task at
a future date.

Consider a question ``Does perturbation $q_p$ result in $q_e$?'' with answer $d_{e} \in \{+,-\}$
(as a shorthand for $\{correct,opposite\}$), created from an IG path:
\vspace{1mm}
\centerline{$q_{p} \xrightarrow{r_{pX}} X \xrightarrow{r_{XY}} Y \xrightarrow{r_{Ye}} q_{e}$} \\
Here, $r_{pX}$, $r_{XY}$, and $r_{Ye}$ denote the polarities (+/-) of
the edges $q_{p}$X, XY, and X$q_{e}$ in the IG respectively. (As described earlier, answer $d_{e}$ is
the product of the polarities $r_{pX}.r_{XY}.r_{Ye}$).
To define an explanation in terms of the paragraph's sentences $x_{1},...,x_{K}$,
we define the gold explanation $E_{gold}$ as the structure:\\
\vspace{1mm}
\centerline{$q_{p} \rightarrow d_i x_i \rightarrow d_j x_j \rightarrow d_e q_e$} \\
where $x_{i}$ is the sentence corresponding to X,
$x_{j}$ is the sentence corresponding to Y,
and $d_{i}$, $d_{j}$, and $d_{e}$ denote the directions of influence (+/-, denoting \{more,correct\}/\{less,opposite\}).
As workers already annotated the alignment between X and $x_{i}$
(similarly Y and $x_{j}$) we know $x_{i}$ and $x_{j}$. Similarly,
as workers also annotated whether X describes an increase of decrease of $x_{i}$, denoted by $d_{X} \in \{+,-\}$,
we can straightforwardly compute the directions of influence:
\begin{myquote}
$d_{i} = r_{pX}.d_{X}$ \\
$d_{j} = d_{i}$ (in-paragraph influence\footnote{Paragraph sentences always describe correct, not opposite, influences on later sentences,
HENCE if $x_{i}$ is more/accelerated, $x_{j}$ will be too (similarly for less/decelerated).}) \\
$d_{e} = r_{pX}.r_{XY}.r_{Ye}$ = the answer 
\end{myquote}
We can similarly generate explanations for answers derived from 1-hop and 2-hop paths.

We generated a full database of explanations for all the questions with answer ``correct'' or ``opposite''
(For ``no effect'' answers, there is no explanation as there is by definition no path of influence).
We then removed the (occasional) explanation
where worker annotations were contradictory (e.g., $j < i$) or had no majority decision for an annotation.
This database is available for a possible future explanation task (given question + paragraph, produce the answer + explanation).

\begin{table}[t]
\begin{mycentering}
{\small
\begin{tabular}{l|ccc|c}
\toprule 
      Count of             & Train & Dev  & Test  & Total \\\midrule
Topics            &  87  &  23 & 12  &   122  \\
Paragraphs        &  261  & 77 & 41   &  379  \\
Influence graphs  & 1453  &  424  &  230 &   2107 \\
Questions         &  29808     &  6894   &  3993    &   40695  \\ 
\bottomrule
\end{tabular}
}
\end{mycentering}

\vspace{4mm}

\resizebox{\columnwidth}{!}{%
\begin{tabular}{ll|rrr|r}
\toprule 
 &       & \multicolumn{4}{c}{\# Questions} \\
 &       & Train & Dev  & Test  & Total \\\midrule
Question & in-para &  7303     &  1655   &    935  &  9893   \\ 
type & out-of-para &  12567     & 2941    & 1598     &    17108 \\ 
         & no-effect & 9936      & 2298    &  1460    &    13694 \\
        &   Total    &  29808   &  6894    & 3993   &  40695 \\\midrule
\midrule
Number  & \#hops=1 &6754 &1510 & 835& 9099   \\ 
of hops & \#hops=2      &8969 &2145 & 1153& 12267    \\ 
(in- \& out-   & \#hops=3     &4149 &941  & 545& 5635   \\ 
    of-para qns)    &   Total  &19872&4596 & 2533& 27001    \\
\bottomrule
\end{tabular}
}
\caption{Dataset statistics}
\label{table:stats_main}
\vspace{-2mm}
\end{table}

\section{Experiments}

\subsection{Models}
We measured the performance of two baselines and three strong neural models on WIQA, to understand how it stresses these models: 
\begin{des}
\item[{\bf Majority}] predicts the most frequent label, {\it correct}, in the training dataset. 
\item[{\bf Polarity}] is a rule-based baseline that assumes influences of the form ``{\it more X $\rightarrow$ more Y}" (similarly for ``less'') are {\it correct},
hence ``{\it more X $\rightarrow$ less Y}'' are {\it opposite}. A small lexicon of positive ({\it ``more''}) and negative ({\it ``less''}) words is used to assign the more/less polarities. A random class label is chosen when assignments cannot be made.

\item[{\bf Adaboost}] \cite{Freund1995ADG} was used to make the 3-way classification using several bag-of-word features
  computed from {\it change} and {\it effect}.

\item[{\bf Decomp-Attn}] applies the 
  Decomposable Attention (DA) model of \cite{parikh-etal-2016-decomposable} to our task.
  The original DA model computes entailment, i.e., the confidence that a {\it premise}
  entails (or contradicts) a {\it hypothesis}. We recast WIQA as an entailment task where cause-effect
  becomes premise-hypothesis, and (correct/opposite/no-effect) correspond to
  (entailment/contradiction/neutral). We retrain the DA model on WIQA using this mapping.
\item[{\bf BERT}] \cite{Devlin2018BERTPO} is a pre-trained transformer-based language model
  that has achieved state of the art performance on many NLP tasks. We supply
  questions to BERT as {\it [CLS] paragraph [SEP] question [SEP] answer-option} for each of the three options.
  The [CLS] token is then projected to a single logit and fed through a softmax layer across the three options,
  using cross entropy loss, and the highest-scoring option selected. We fine-tune BERT on
  the WIQA training data in this way. We also measure
  an ablated version where the paragraph is omitted (train and test).
\item[{\bf Human Performance}] was estimated by
  having three people independently answering the same 100 questions (with paragraphs) drawn randomly from the test set. Krippendorff's alpha (nominal metric) for these answers was 0.908 (high agreement) \cite{krippendorff1970estimating}.
  
\end{des}

\vspace{-1mm}
\section{Results and Analysis}
\vspace{-1mm}
\subsection{Prediction Accuracy}

\begin{table}[t]
\resizebox{\columnwidth}{!}{%
\setlength\tabcolsep{4pt}  
\begin{tabular}{l|ccc|c}
\toprule
   Question Type             & in-para  & out-of-para & no-effect & Total \\ \midrule
\# questions             &935 &1598 &1460 &3993 \\\midrule
\emph{Majority}               &45.46 &49.47 &0.55 &30.66 \\
\emph{Polarity}               &76.31 &53.59 &0.27 &39.43 \\
\emph{Adaboost}                & 49.41 & 36.61  &48.42  &43.93   \\
\emph{Decomp-Attn}             &56.31 &48.56 &73.42 & 59.48\\
\emph{BERT (no para)}    &60.32 &43.74 &84.18 &62.41 \\
\emph{BERT} &\textbf{79.68} &\textbf{56.13} &\textbf{89.38} &\textbf{73.80} \\
\hline
 Human perf.      &  & &  & 96.33\\
\bottomrule
\end{tabular}}
\caption{Comparing models on \ourdata{} test partition }
\label{table:results_wiqa_main}
\vspace{-2mm}
\end{table}

\noindent
The results (Table~\ref{table:results_wiqa_main}) provide several insights:

\noindent
1. {\bf The dataset is hard.} Our strongest model (73.8) is over 20 points behind human performance, suggesting WIQA poses significant challenges. Prediction of out-of-para effects is particularly challenging, 37 points behind human performance.

\noindent
2. {\bf BERT already ``knows'' some change-effect knowledge.} Even without the paragraph,
and even though the test paragraphs are on topics unseen in training, BERT scores substantially above the baselines.
This suggests BERT has some type of cause-effect information embedded in it.

\noindent
3. {\bf Supplying the paragraph helps,} resulting in 10 points higher score, illustrating that WIQA contains questions that require
understanding of the paragraph. This suggests more sophisticated reading strategies may further improve results.

\begin{figure}[t]
  \includegraphics[width=0.45\textwidth]{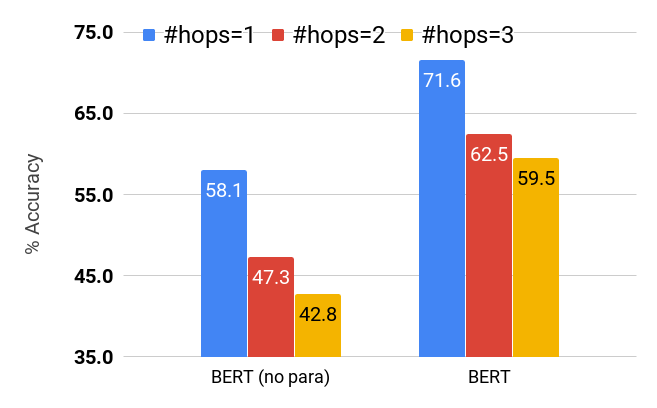}
  \caption{Accuracy of the best baselines drops as number of hops increase, quicker for `no para' version.
  }
  \label{fig:vary_chain_length}
\end{figure}

\subsection{Predicting Indirect Effects}

In-para and out-of-para questions were derived from chains of different lengths (``hops'') in the influence graphs.
Figure~\ref{fig:vary_chain_length} shows how performance varies with respect to those lengths,
and shows that {\bf ``indirect'' (2/3-hop) effects are harder to predict than ``direct'' (1-hop) effects}.
For example, it is easier to predict ``cloudy day'' results in ``less sunshine'' (direct)
than ``less photosynthesis'' (indirect). This suggests that some form of reasoning along influence chains
may be needed to predict indirect effects reliably, as those effects are less likely to be 
explicitly stated in corpora and embedded in pre-trained language models.


\subsection{Consistency}

\begin{figure}[t]
  \includegraphics[width=0.45\textwidth]{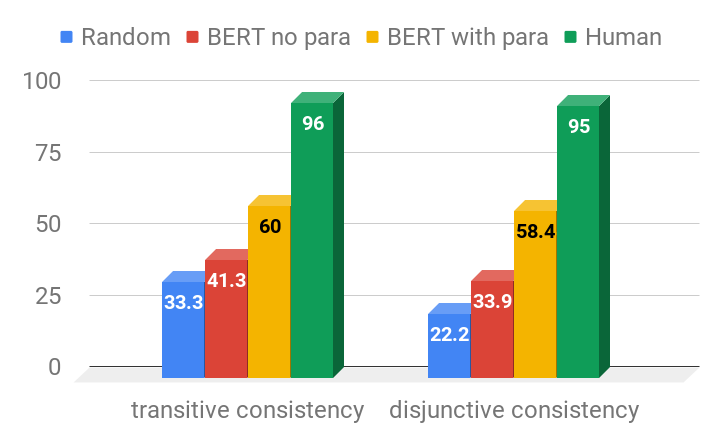}
  \caption{The best models (red,yellow) make substantially less consistent predictions than humans (green).
    }
  \label{fig:consistency-graph}
\end{figure}

Are the models making consistent predictions?
If a model predicts both X$\rightarrow$Y and Y$\rightarrow$Z are correct, it
should, if it were consistent, also predict $X\rightarrow$Z is correct.
To measure a model's {\it transitivity consistency}, for each influence graph, we
measure how often its indirect predictions (2/3-hop) are consistent\footnote{
i.e., the polarity (+/-, for correct/opposite) of edge XZ = the product of the polarities of edges chaining from X to Z. As models can also 
predict ``no effect'', random score is 1/3.}
with its 1-hop predictions. Similarly, we measure {\it disjunctive consistency}
by how often its predictions for edges known to be opposite (eg
X$\rightarrow$Y and X$\rightarrow${\it opp-effect-in-para}
in Fig~\ref{fig:consistency})
are indeed so\footnote{
Only edge pairs with labels +\&-, or -\&+, are disjunctively consistent (of 9 possible labelings), hence random is 2/9. 
}. The results in Figure~\ref{fig:consistency-graph} illustrate that the {\bf models are
far from consistent}. This suggests that reasoning with global consistency
constraints may improve results, e.g., \cite{ning2017structured,propara-emnlp18}.


\subsection{Linguistic and Semantic Phenomena}

We analyzed 200 descriptions of changes in 100 random questions,
and observe the following challenging (overlapping) phenomena to handle:

\noindent
{\bf 1. Qualitative Language:} $\approx$65\% of the change statements 
are expressed qualitatively, using a
broad vocabulary of comparatives (e.g., more, fewer, smaller, larger,
cooler, slower, higher, harder, decreased, hotter) or their corresponding
adjectives (small, cool, etc.). In addition, whether the change is
a positive or negative influence on the process is context-dependent
(``more X'' can be positive or negative, depending on X, and
sometimes depending on the paragraph topic itself).

\noindent
{\bf 2. Commonsense} ($\approx$45\%):
Exogenous influences are (by definition) not stated in
the paragraph, and so require substantial commonsense to
understand, e.g., that ``heavy rainfall'' (out of para)
negatively influences ``more wild fires'' (in para);
or that ``overfishing'' (out of para) negatively
influences ``fish lay eggs'' (in para).

\noindent
{\bf 3. Lexical matching} $\approx$15\%
of the in-para changes refer to paragraph
entities using different terms, e.g.,
``insect'' (para) $\leftrightarrow$ ``bee'' (question),
``becomes'' $\leftrightarrow$ ``forms'',
``removes'' $\leftrightarrow$ ``expels'',
complicating aligning questions with the paragraph.

\noindent
{\bf 4. Negation} ($\approx$6\%):
Negation occurs in about 6\% of the changes, e.g., ``drought does not
occur'', ``soil is not fertile'', ``magma does not get larger''.

\noindent
{\bf 5. Juxtaposed polarities} ($\approx$3\%):
Sometimes positive- and negative-related terms are juxtaposed,
(e.g., ``much less'', ``increased deforestation'', ``less severe'') again
challenging to process.

\noindent
These all illustrate the diversity of linguistic and semantic challenges in WIQA.

\section{Conclusion}

An important test of understanding procedural text is
whether the effects of {\it perturbations} to the process
can be predicted. To that end, we have introduced WIQA,
the first large-scale dataset for ``what if'' reasoning over text.
While our experiments suggest language models have some built-in
knowledge of influences, and some ability to identify influences
in paragraphs, these capabilities are limited, producing
predictions that are over 20 points worse than humans,
often inconsistent, and particularly erroneous about
indirect (multi-hop) effects.
WIQA aims to improve this state of affairs, offering a new challenge and
resource to the community.
The dataset is available at http://data.allenai.org/wiqa/.

\subsection*{Acknowledgements}
We are grateful to the AllenNLP and Beaker teams at AI2, and for the insightful discussions with other Aristo team members. 
Computations on beaker.org were supported in part by credits from Google Cloud.

\bibliography{references}

\bibliographystyle{acl_natbib}

\appendix

\section*{Appendix A: Topicwise consistency}
We study trends in topic-wise accuracy of models as they read more context information. Bert no-para model does not have access to any context or paragraph, except the language model's background knowledge from Wikipedia. By reading the paragraph context Bert with-para model performs much better on certain topics such as \texttt{Pollination, blood, mountain, evaporation} but the impact of reading is much less on topics such as \texttt{Igneous rocks, plant crops, solar eclipse, DNA replication}. Topics such as \texttt{blood} are very popular on Wikipedia and distributed across several very different articles. These topics are harder for BERT as it requires additional paragraph context to understand the question.

\begin{table}[h]
\begin{tabular}{|l|l|l|}
\hline
topic           & BERT (no para) & BERT \\
\hline
igneous rock    & 0.66           & 0.64           \\
plant crops     & 0.61           & 0.61           \\
solar eclipse   & 0.43           & 0.43           \\
frog            & 0.59           & 0.62           \\
DNA replication & 0.58           & 0.63           \\
water cycle     & 0.63           & 0.69           \\
fish            & 0.5            & 0.57           \\
pumpkin         & 0.61           & 0.69           \\
pollination     & 0.62           & 0.75           \\
blood           & 0.62           & 0.76           \\
mountain        & 0.57           & 0.72           \\
evaporation     & 0.42           & 0.67  \\ 
\bottomrule
\end{tabular}
\caption{As the Bert model (that has access to the paragraph in context) reads more paragraphs in context, its accuracy is better. Reading helps certain topics such as \texttt{Pollination, blood, mountain, evaporation} more than others}
\end{table}

\section*{Appendix B: Crowdsourcing Influence Graphs}
\begin{figure}[!h]
  \includegraphics[width=0.5\textwidth]{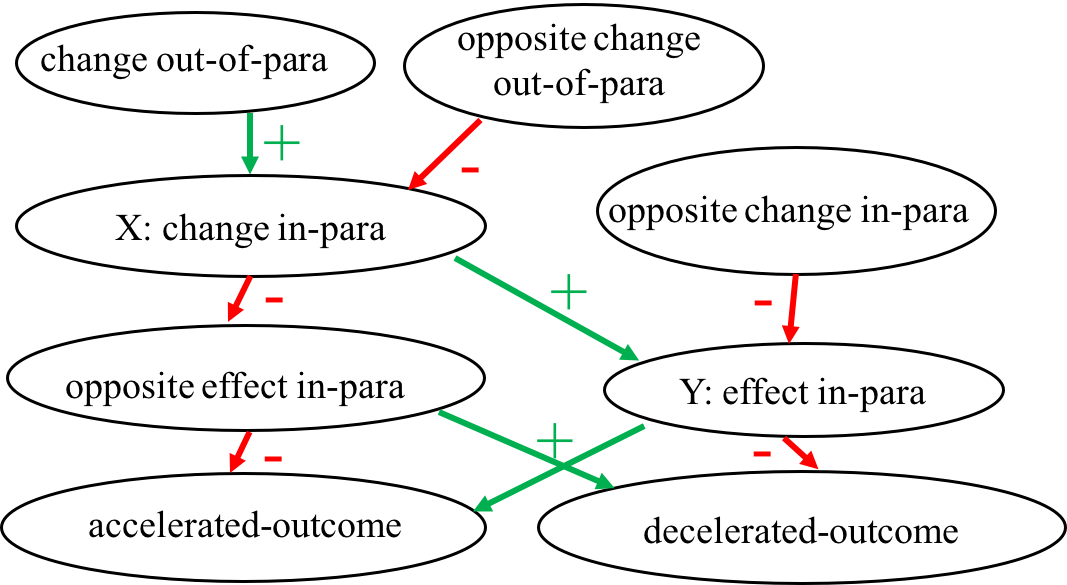}
  \caption{At the start of the process to annotate an influence graph for a given paragraph, the annotators see a blank influence graph with the basic structure.}
  \label{fig:blank_influence_graph}
\end{figure}

\begin{figure*}[!h]
  \includegraphics[width=1.0\textwidth]{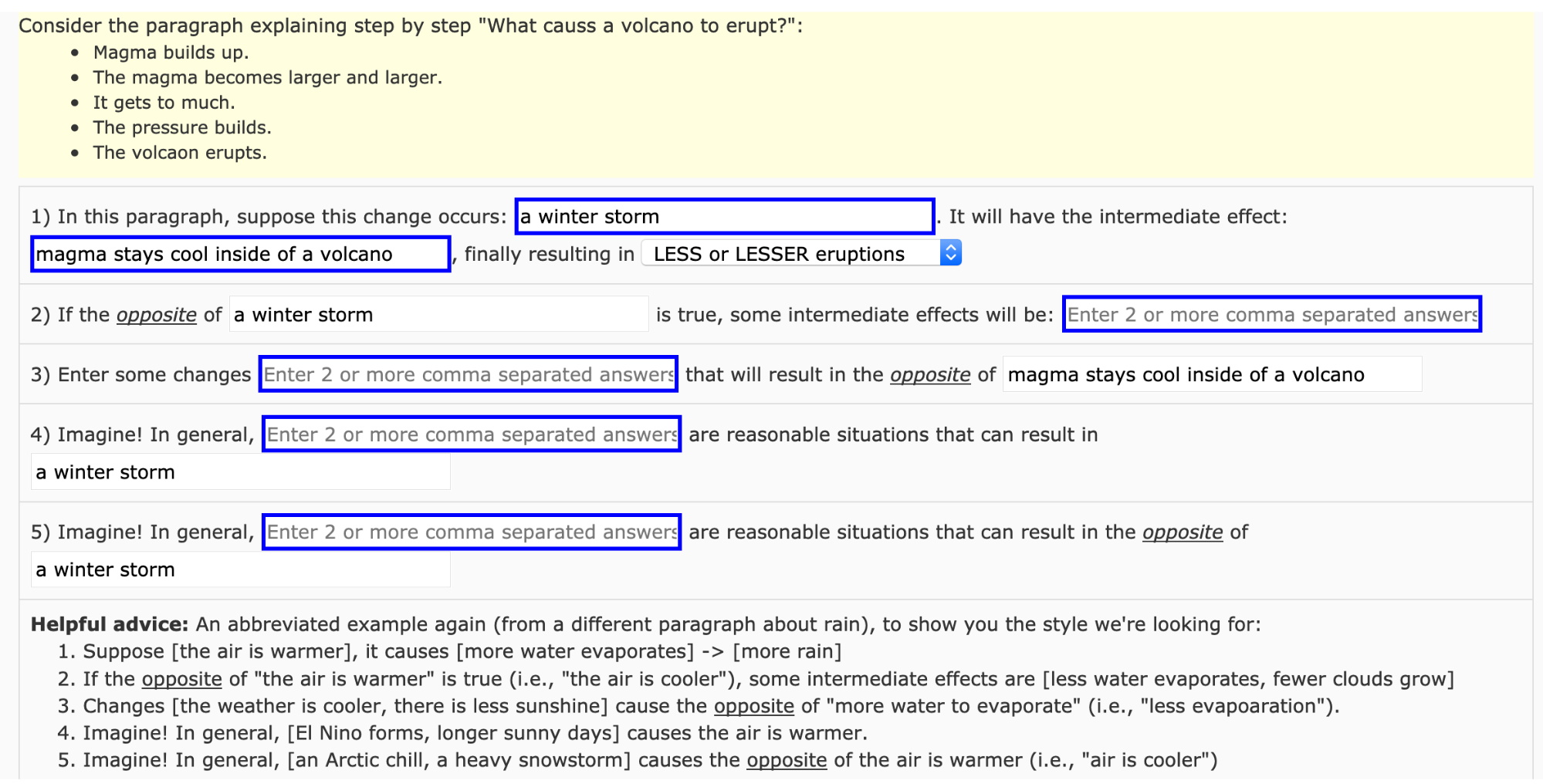}
  \caption{The interface shown to the annotators on Mechanical turk platform. Given a paragraph in yellow background, the annotators answer the five questions and an influence graph emerges from their answers.}
  \label{fig:mturk_template}
\end{figure*}

We crowdsource influence graphs by getting the graphs constructed progressively, with the help of five questions stated in Figure \ref{fig:mturk_template}. At first, the turkers see an empty graph in Figure \ref{fig:blank_influence_graph}. 

\begin{figure}[!h]
  \includegraphics[width=0.5\textwidth]{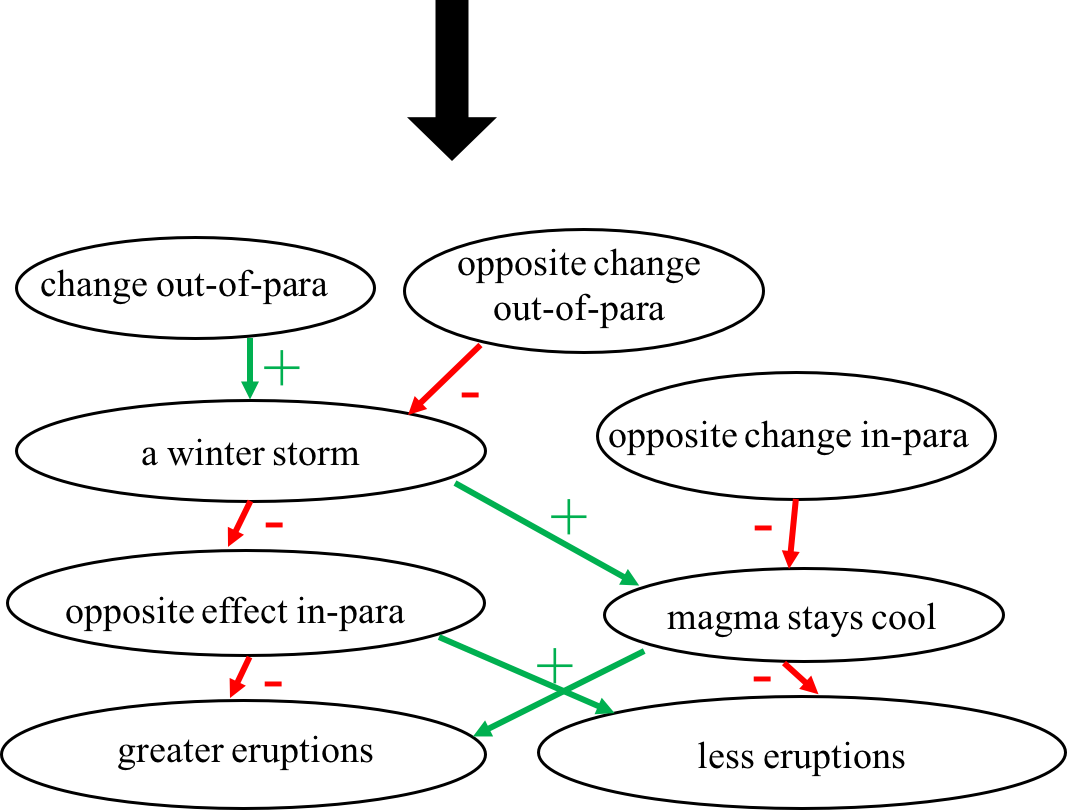}
  \caption{As the annotators answer questions in Fig. \ref{fig:mturk_template}, a partial influence graph emerges. As they answer questions, the annotators found it useful to validate their answers by examining the emerged influence graph.}
  \label{fig:partial_influence_graph}
\end{figure}

When the annotators answer the first question (shown in Fig. \ref{fig:mturk_template}), two nodes of the partial influence graph are filled (depicted in Fig. \ref{fig:partial_influence_graph}).

Once all the questions are answered, the influence graph will be ready. During the process of annotation, there are appropriate validations for quality control.

\section*{Appendix C: Sample Influence Graphs}
\begin{figure*}[t]
  \includegraphics[width=1\textwidth]{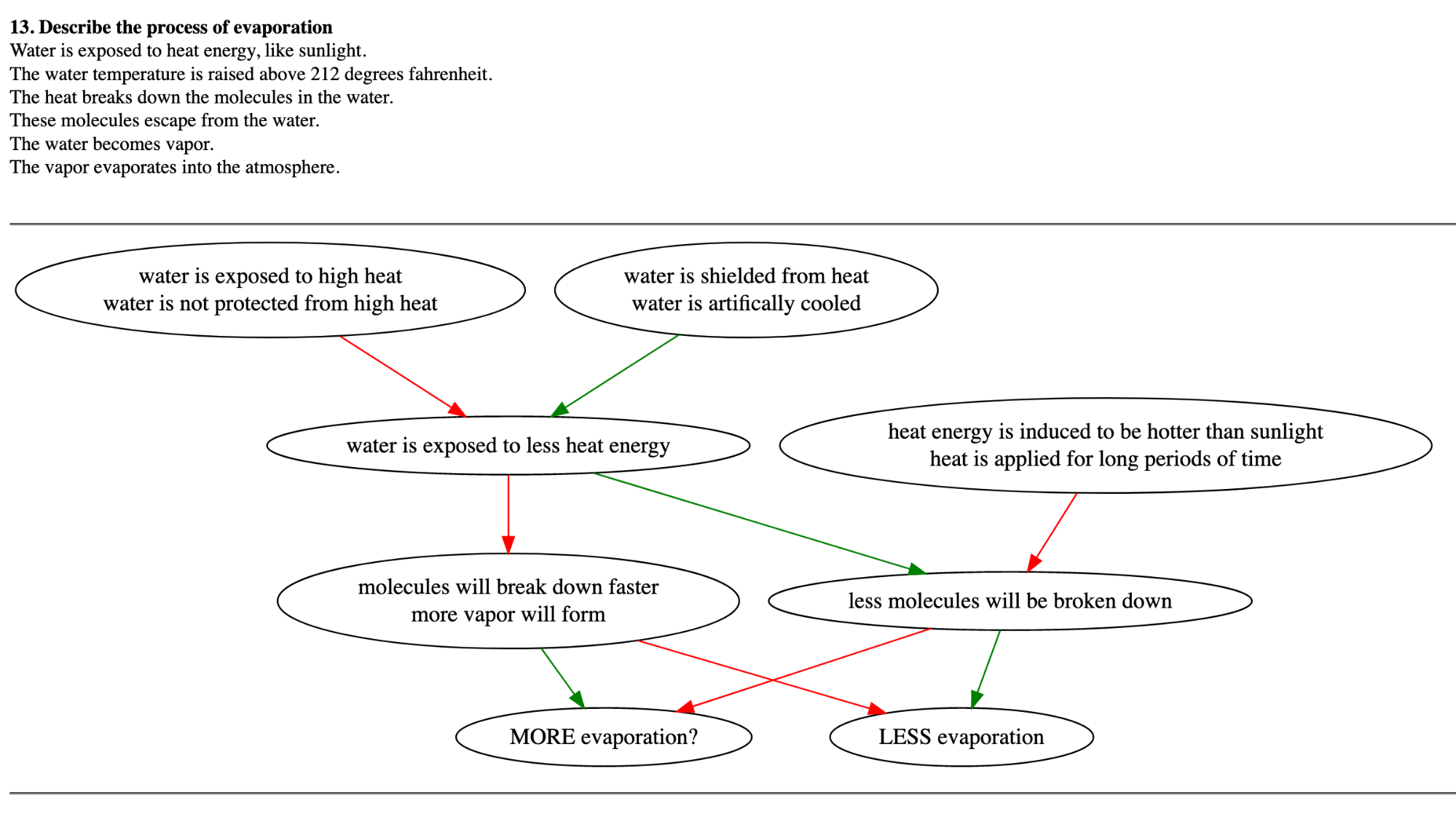}
  \caption{Influence graph for a paragraph from the topic \texttt{evaporation}}
  
  \label{fig:example_influence_graph_1}
\end{figure*}

\begin{figure*}[t]
  \includegraphics[width=1\textwidth]{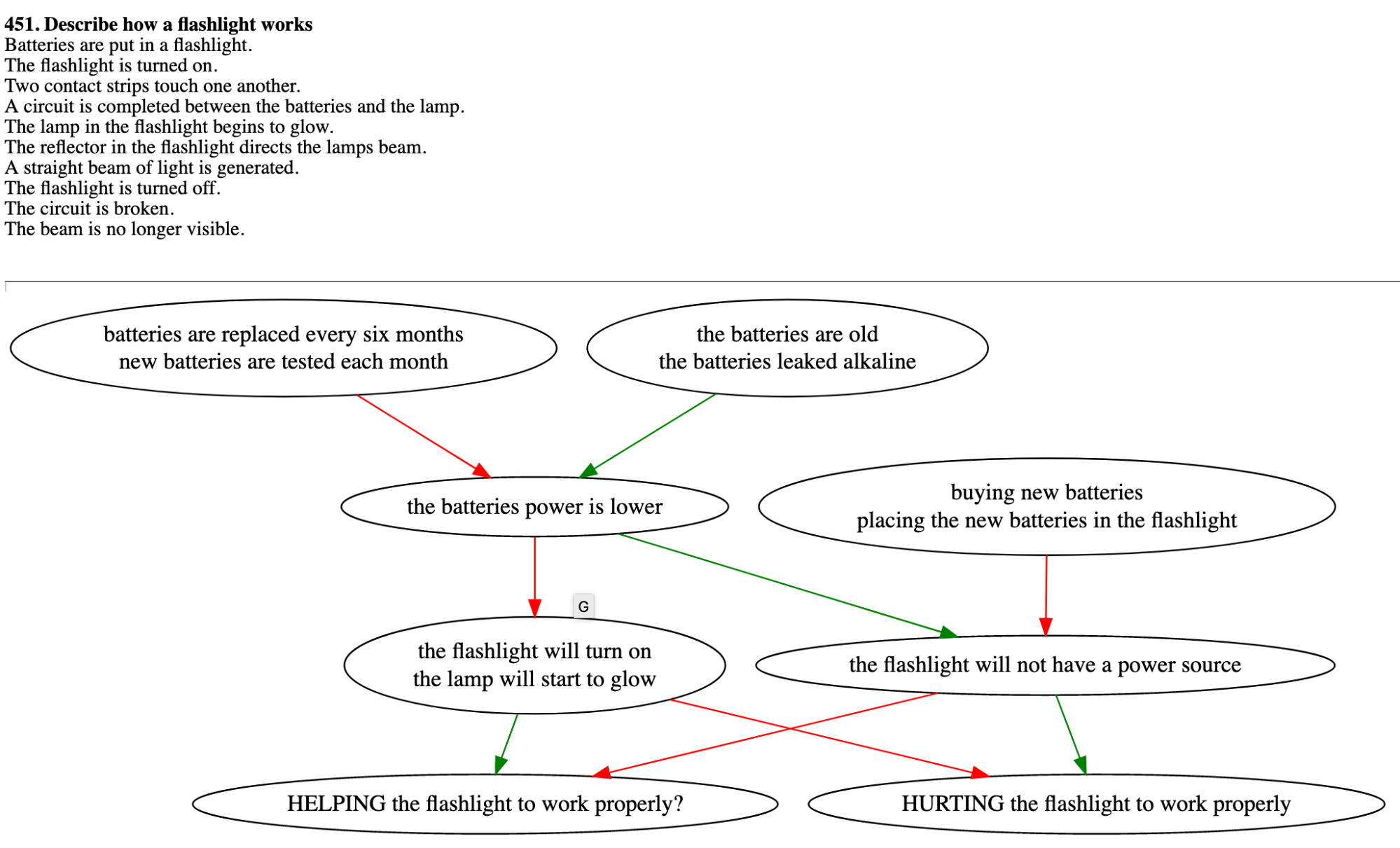}
  \caption{Influence graph for a paragraph from the topic \texttt{flashlight}}
  
  \label{fig:example_influence_graph_2}
\end{figure*}

\begin{figure*}[t]
  \includegraphics[width=1\textwidth]{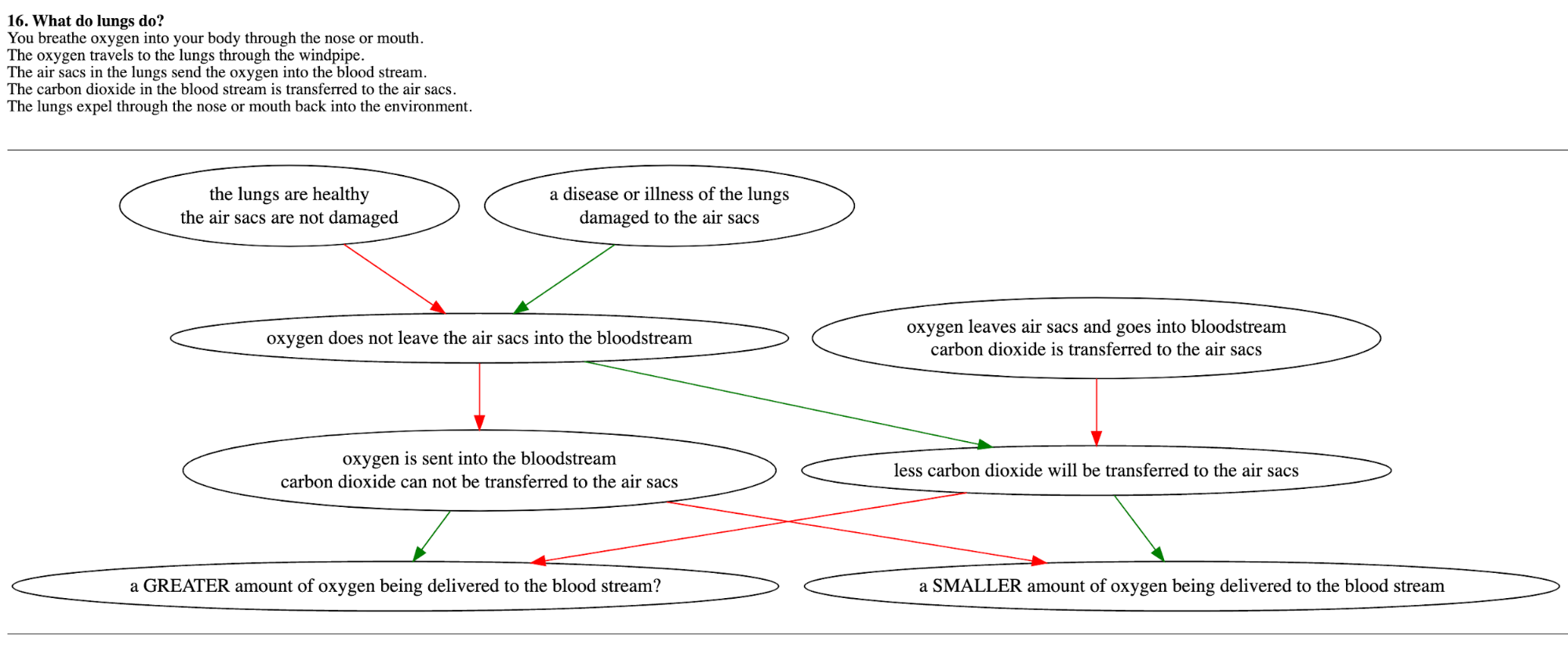}
  \caption{Influence graph for a paragraph from the topic \texttt{lungs}}
  
  \label{fig:example_influence_graph_3}
\end{figure*}

\begin{figure*}[t]
  \includegraphics[width=1\textwidth]{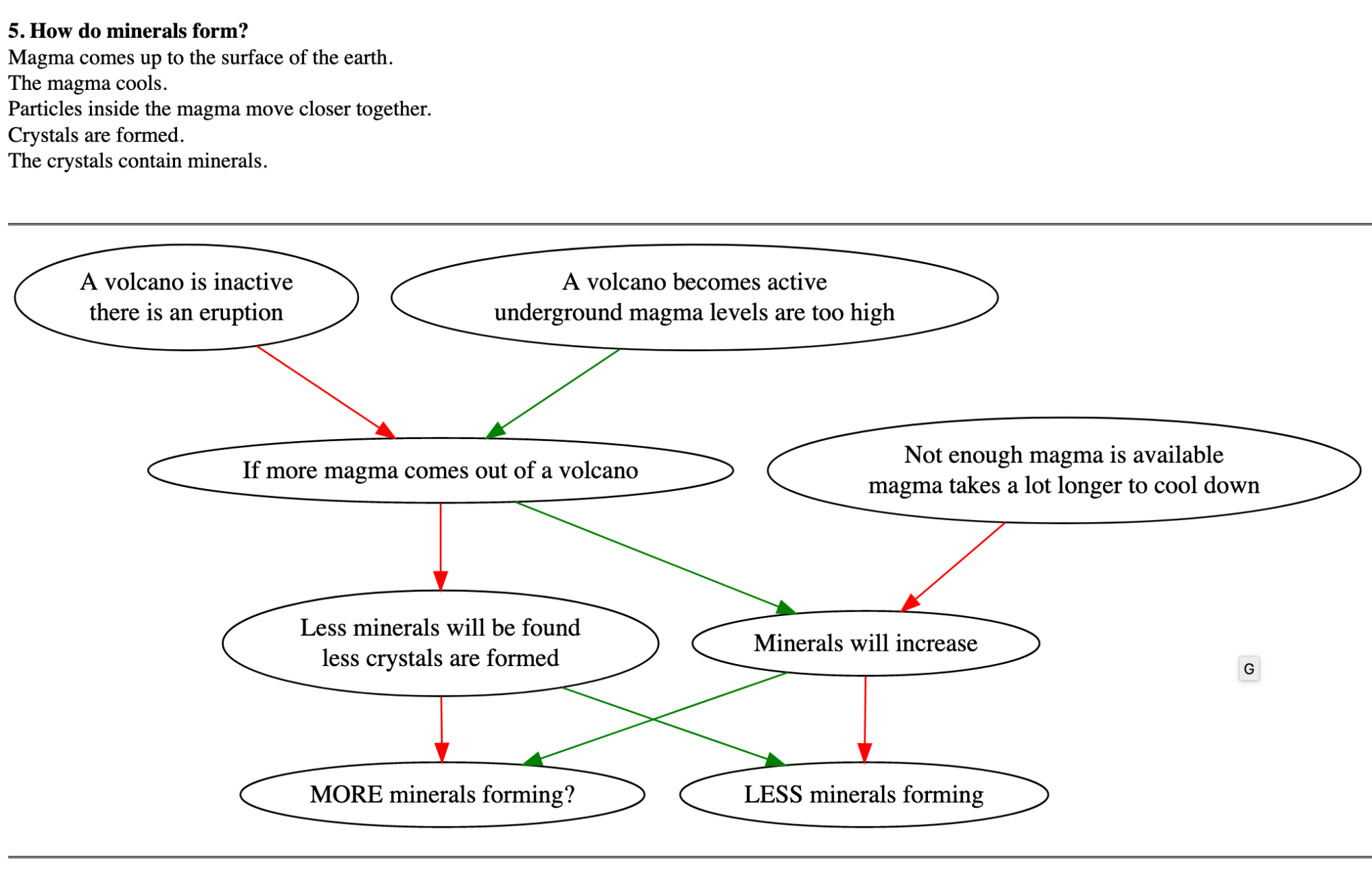}
  \caption{Influence graph for a paragraph from the topic \texttt{minerals}}
  
  \label{fig:example_influence_graph_4}
\end{figure*}

To get an impression of our crowdsourced influence graph repository, we display four paragraphs (not hand picked) in Figures \ref{fig:example_influence_graph_1}, \ref{fig:example_influence_graph_2},
\ref{fig:example_influence_graph_3},
\ref{fig:example_influence_graph_4}. These range from natural process, to human body process and mechanical process. 




\end{document}